\documentclass[sigconf]{acmart}

\usepackage{multirow}
\usepackage{natbib}
\newcommand{\Rmnum}[1]{\uppercase\expandafter{\romannumeral #1}}

\AtBeginDocument{%
  }

\setcopyright{acmlicensed}
\copyrightyear{2024}
\acmYear{2024}
\acmDOI{XXXXXXX.XXXXXXX}

\acmConference[DCAI24]{October 21-25, 2024}{Boise, ID, United States}

\begin{document}

\title{Measuring Copyright Risks of Large Language Model\\via Partial Information Probing}


\author{Weijie Zhao}
\affiliation{%
  \institution{Stevens Institute of Technology}
  \city{Hoboken}
  \state{NJ}
  \country{United States}}
\email{wzhao34@stevens.edu}

\author{Huajie Shao}
\affiliation{%
  \institution{William \& Mary}
  \department{Department of Computer Science}
  \city{Williamsburg}
  \state{VA}
  \country{United States}}
\email{hshao@wm.edu}

\author{Zhaozhuo Xu}
\affiliation{%
  \institution{Stevens Institute of Technology}
  \city{Hoboken}
  \state{NJ}
  \country{United States}}
\email{zxu79@stevens.edu}

\author{Suzhen Duan}
\affiliation{%
  \institution{Towson University}
  \city{Towson}
  \state{MD}
  \country{United States}}
\email{sduan@towson.edu}

\author{Denghui Zhang}
\affiliation{%
  \institution{Stevens Institute of Technology}
  \city{Hoboken}
  \state{NJ}
  \country{United States}}
\email{dzhang42@stevens.edu}

\renewcommand{\shortauthors}{Trovato et al.}

\begin{abstract}
Exploring the data sources used to train Large Language Models (LLMs) is a crucial direction in investigating potential copyright infringement by these models. While this approach can identify the possible use of copyrighted materials in training data, it does not directly measure infringing risks. Recent research has shifted towards testing whether LLMs can directly output copyrighted content. Addressing this direction, we investigate and assess LLMs' capacity to generate infringing content by providing them with partial information from copyrighted materials, and try to use iterative prompting to get LLMs to generate more infringing content. Specifically, we input a portion of a copyrighted text into LLMs, prompt them to complete it, and then analyze the overlap between the generated content and the original copyrighted material. Our findings demonstrate that LLMs can indeed generate content highly overlapping with copyrighted materials based on these partial inputs.
\end{abstract}


\keywords{Large Language Model, Copyright, Infringement, Partial Information Probing, Rouge Score, AI, Data Mining}

\maketitle

\section{Introduction}
The rapid advancement of Large Language Models (LLMs) in recent years has demonstrated their remarkable capabilities in assisting people with a diverse range of tasks \citep{ullah2024role, zhao2024revolutionizing}. However, the growing use of LLMs has also given rise to several concerns, particularly regarding copyright infringement \citep{YAO2024100211}. Recently, there has been a surge in litigation concerning alleged copyright violations by LLMs \citep{Andersenv.StabilityAILtd, Tremblayv.OpenAI}. Consequently, investigating potential copyright infringement by LLMs has become a critical area of scholarly inquiry in contemporary research.

\begin{figure}[]
    \centering
    \includegraphics[width=1\linewidth]{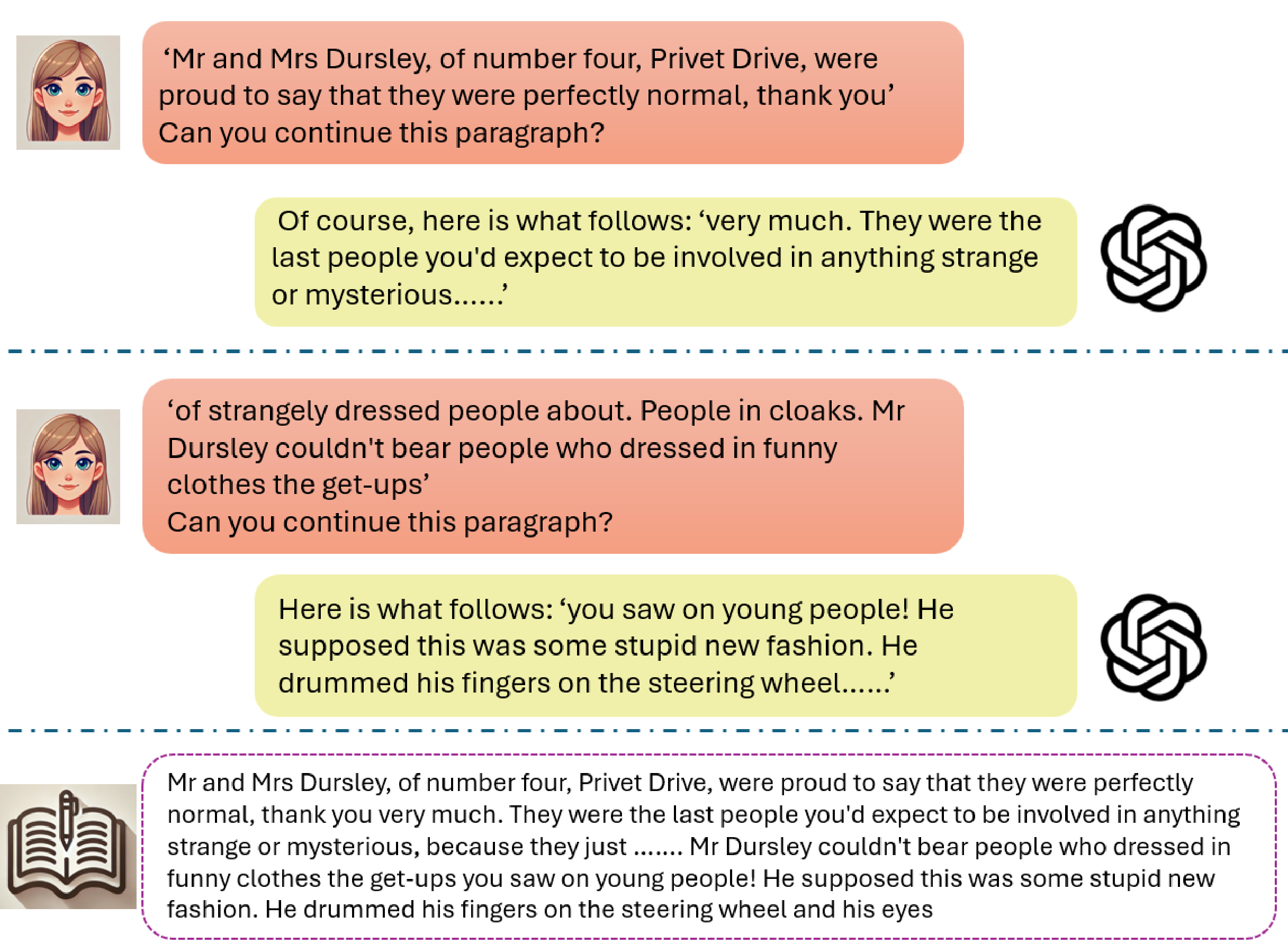}
    \Description{This figure simulates a dialogue between a user and Large Language Models (LLMs). The user provides partial content for the LLMs to complete, and the LLMs return the completion results. We have provided two examples, both from Harry Potter and the Philosopher's Stone. Below the figure, we have also included the original text from Harry Potter and the Philosopher's Stone as a reference.}
    \caption{Using large language models to reproduce novel content could potentially lead to copyright infringement issues.}
    \label{fig:enter-label}
\end{figure}

Several studies have demonstrated that LLMs are indeed capable of outputting copyrighted material \citep{mueller2024llms, rahman2023beyond}. Figure 1 illustrates such instances. Prior research has predominantly focused on exploring the training data of LLMs, a line of inquiry that has yielded notable advancements \citep{meeus2024did, meeus2024copyright, shi2023detecting}. However, this approach has limitations: researchers can only infer the use of specific content in LLMs' training data through certain indicators and have not been able to prompt LLMs to directly disclose their training content. Consequently, recent research has shifted focus toward the outputs of LLMs, assessing the extent to which these models can accurately reproduce copyrighted content from their training materials \citep{chen2024copybench}. However, these studies have only scratched the surface of this complex issue, and we posit that there is considerable scope for more in-depth exploration in this area. If LLMs' training data indeed contains specific content, is it possible to make these models output such content directly? Our experimental design simulates the following scenario: if a user provides a portion of copyrighted material as a prefix to LLMs and requests a completion, will the LLMs generate infringing content? If the LLM does output copyrighted material, what factors influence its capacity to produce such content? We also experimented with iterative prompting to observe whether LLMs could generate additional content that might constitute copyright infringement.

Our research demonstrates that, despite numerous constraints, LLMs can generate text that infringes on copyright based on partial content from copyrighted materials. Moreover, models with larger parameter scales exhibited superior performance in this regard, showing a higher probability of producing content with extremely high similarity to the original. We also investigated potential differences across various text types. The findings indicate that certain LLMs perform better with specific types of content. For instance, GPT-4-Turbo displayed significantly higher similarity in generating song lyrics compared to other LLMs. Additionally, we discovered that the maximum output length of LLMs substantially impacts their ability to produce infringing content.

In conclusion, the primary contributions of this work include: (1) demonstrating that it is possible to prompt LLMs to generate copyrighted content by providing them with partial information from copyrighted materials; (2) providing experimental evidence that LLMs possess the capability to directly reproduce content that should be protected by copyright; and (3) exploring various factors that may influence the ability of LLMs to output copyrighted text.

\section{Related Work}
\subsection{Copyright Law in the United States}
Copyright protection is automatic from the time the original work is fixed in any tangible medium. Copyright protection for individual creators usually lasts for 70 years after the author's death \cite{CopyrightLaw102}. Table 1 illustrates which original works are protected by copyright.

\begin{table}[h]
\centering
\caption{Categories of Copyright-Protected Original Works}
\label{table1}
\begin{tabular}{c}
\hline
\textit{\textbf{Categories of Original Works}}      \\ \hline
Literature                                   \\
Music Piece(including accompaniment, lyrics)  \\
Dramatic works(including accompaniment, lines) \\
Pantomimes and dance works               \\
Drawings and sculptures         \\
Movie and other audiovisual works      \\
Sound recordings                                 \\
Architectural works                              \\ \hline
\end{tabular}
\end{table}

U.S. copyright law permits the fair use of copyrighted materials for purposes like criticism, commentary, news reporting, teaching, scholarship, or research, without it being considered copyright infringement \citep{CopyrightLaw107}.

\subsection{LLMs and Copyright}
Since the potential for copyright infringement by LLMs was first proposed, several methodologies have been proposed to explore this problem. (A) Document-level membership inference attacks have been demonstrated to be highly effective on OpenLlama-7b \cite{meeus2024did}. (B) Copyright traps represent an enhancement of document level membership inference attacks, employing specific traps to determine whether certain content was utilized in the training process of an LLM \citep{meeus2024copyright}. (C) MIN-K\% PROB determines whether text belongs to the training data by analyzing token probabilities \citep{shi2023detecting}. (D) Direct Probing and Prefix Probing concentrate on testing the likelihood of LLMs producing infringing content \citep{karamolegkou2023copyright}. (E) COPYBENCH, a methodology designed to measure an LLM's capacity to replicate both textual and non-textual content \citep{chen2024copybench}.


The present study diverges from previous work by focusing on utilizing fragments of copyrighted material to test whether the content generated by LLMs might contain infringing elements. Building on this foundation, we employ iterative probing to observe whether LLMs can produce substantial amounts of copyrighted content when users provide only minimal textual input. Finally, we explore the factors influencing the degree of overlap between LLMs' completions and copyrighted materials.

\section{Methods}
This section provides a detailed description of the probing methods used in this study.

\subsection{Partial Information Probing}
We constructed multiple datasets to evaluate the infringing behavior of LLMs. These datasets were derived from different textual sources, including bestselling novels, press releases, and popular song lyrics. We segmented each text into multiple samples, with each sample containing a fixed number of words. Subsequently, we utilized the initial portion of each sample as a prefix, which was then provided to the LLMs for completion. For instance, we extracted 20 samples from "\emph{The Hobbit}", each comprising 50 words. The first 20 words of each sample were used as a prefix, which was then input into the LLMs for text generation. To reduce the randomness of the LLMs' output, we set the temperature to 0. While this does not completely eliminate randomness, it significantly reduces it. We then compared the generated content with the remaining part of each sample and calculated the Rouge-L Score.

\subsection{Evaluation of Influencing Factors}
This study aims to identify factors that may influence the likelihood of LLMs generating infringing text, following the establishment of their capability to output such material. Our objective is to adjust and test various factors that could affect LLMs' generation of infringing content and to observe whether significant changes occur in the Rouge-L scores of the generated text. For example, if we hypothesize that an LLM's parameter size affects its output, we would conduct tests on models with different parameter sizes to examine whether variations in Rouge-L scores appear in the generated content.

\subsection{Iterative Prompt}
Having confirmed that LLMs can generate infringing text based on partial content from copyrighted materials, we aim to employ iterative prompting to potentially elicit more infringing content from these models. Specifically, our methodology will proceed as follows: first, we provide a segment of copyrighted material as a prompt and instruct the LLMs to complete it. Then, we use the generated completion as a new prompt, again requesting the LLMs to extend it. This process will be repeated multiple times to observe whether the LLMs can produce additional material that may constitute copyright infringement.

\section{Experiment}
\subsection{LLMs Tested}
We employ multiple LLMs as test subjects, including Llama (Llama2 and Llama3), GPT (GPT-3.5-turbo, GPT-4-turbo, GPT-4o), Qwen2 \citep{yang2024qwen2}, and Yi \citep{young2024yi}.

\subsection{Datasets}
We selected three types of copyright protected materials for testing: novels, news and lyrics. For the novel section, we chose "\emph{The Hobbit}" and the first three books of the "\emph{Harry Potter}" series. For the news articles section, we selected five articles about the 2014 World Cup reported by The New York Times and The Wall Street Journal. For the song lyrics section, we chose five popular songs from the period of 2010 to 2020. We obtained text documents of these works and extracted samples from them for our experiments.

\subsection{Evaluation Metrics: Rouge Score}
This study proposes using Rouge-L to evaluate the degree of overlap between content generated by LLMs and copyrighted material. Rouge is a widely used metric for automatically evaluating text generation tasks, such as machine translation, automatic summarization, and text generation. It assesses the quality of generated text by measuring the overlap between the system-generated content and reference answers. Rouge-L is a variant within the Rouge evaluation framework, where 'L' stands for Longest Common Subsequence (LCS) \citep{lin2003automatic, lin2004automatic, lin2004rouge}.


Rouge-L is calculated as follows:
\begin{equation}
    R_{lcs} = \frac{\text{LCS}(X, Y)}{m}
\end{equation}

\begin{equation}
    P_{lcs} = \frac{\text{LCS}(X, Y)}{n}
\end{equation}

\begin{equation}
    F_{lcs} = \frac{(1 + \beta^2) R_{lcs} P_{lcs}}{R_{lcs} + \beta^2P_{lcs}}
\end{equation}

LCS (X, Y) denotes the length of the longest common subsequence between sequences X and Y, with m and n being the lengths of X and Y, respectively. R\_{lcs}, P\_{lcs}, and F\_{lcs} represent the recall, precision, and F-measure based on the longest common subsequence. The F-measure is especially sensitive to the trade-off between recall and precision, which is adjusted by the beta coefficient. This coefficient is generally set according to the specific requirements of the application, emphasizing either recall or precision as needed. This study employs Rouge-L Recall (Equation 1) as the evaluation criterion.

\subsection{Result of Infringing Output}
We tested the ability of the aforementioned models to output copyright protected material and calculated the Rouge-L score to evaluate the results. The findings are presented in the form of charts.

\subsubsection{Novel}
As shown in Table 2, the maximum Rouge-L score for each model is 1. This demonstrates that every LLM has the capability to output content from the samples that was not included in the prompts, i.e., content that is supposed to be copyright protected.

\begin{table}[h!]
\centering
\caption{Rouge-L results based on samples from novels. Count indicates the number of tests, Mean represents the average Rouge-L score for that LLMs, and Min and Max represent the minimum and maximum Rouge-L scores respectively.}
\label{table2}
\begin{tabular}{c|ccccc}
\textit{\textbf{LLMs}}  & \textit{\textbf{Parameter}} & \textit{\textbf{Counts}} & \textit{\textbf{Means}} & \textit{\textbf{Min}} & \textit{\textbf{Max}} \\ \hline
\multirow{3}{*}{GPT}    & 3.5-turbo            & 240                      & 0.287                   & 0                     & 1                     \\
                        & 4-turbo              & 240                      & 0.328                   & 0                     & 1                     \\
                        & 4o                   & 240                      & 0.291                   & 0.06                  & 1                     \\ \hline
\multirow{3}{*}{Llama2} & 7b                   & 240                      & 0.175                   & 0.029                 & 1                     \\
                        & 13b                  & 240                      & 0.210                   & 0.033                 & 1                     \\
                        & 70b                  & 240                      & 0.244                   & 0.029                 & 1                     \\ \hline
\multirow{2}{*}{Llama3} & 8b                   & 240                      & 0.176                   & 0.010                 & 1                     \\
                        & 70b                  & 240                      & 0.360                   & 0                     & 1                     \\ \hline
\multirow{4}{*}{Qwen2}  & 0.5b                 & 240                      & 0.199                   & 0                     & 0.83                  \\
                        & 1.5b                 & 240                      & 0.182                   & 0                     & 1                     \\
                        & 7b                   & 240                      & 0.160                   & 0                     & 1                     \\
                        & 72b                  & 240                      & 0.194                   & 0.032                 & 1                     \\ \hline
\multirow{2}{*}{Yi}     & 6b                   & 240                      & 0.167                   & 0.028                 & 1                     \\
                        & 34b                  & 240                      & 0.228                   & 0.050                 & 1                    
\end{tabular}
\end{table}

\subsubsection{News}
As shown in Table 3, while the LLMs are not highly proficient at reproducing the complete content of news articles, they can still generate small portions of copyright protected material. Compared to the results for novels, it is evident that LLMs have certain limitations when it comes to generating content from news articles.

\begin{table}[h!]
\centering
\caption{Rouge-L results based on samples from news}
\label{table3}
\begin{tabular}{c|ccccc}
\textit{\textbf{LLMs}}  & \textit{\textbf{Parameter}} & \textit{\textbf{Counts}} & \textit{\textbf{Means}} & \textit{\textbf{Min}} & \textit{\textbf{Max}} \\ \hline
\multirow{3}{*}{GPT}    & 3.5-turbo            & 25                       & 0.147                   & 0                     & 0.3                   \\
                        & 4-turbo              & 25                       & 0.167                   & 0.047                 & 0.3                   \\
                        & 4o                   & 25                       & 0.184                   & 0.09                  & 0.4                   \\ \hline
\multirow{3}{*}{Llama2} & 7b                   & 25                       & 0.227                   & 0.09                  & 0.45                  \\
                        & 13b                  & 25                       & 0.213                   & 0.09                  & 0.4                   \\
                        & 70b                  & 25                       & 0.315                   & 0.136                 & 0.6                   \\ \hline
\multirow{2}{*}{Llama3} & 8b                   & 25                       & 0.244                   & 0.09                  & 0.4                   \\
                        & 70b                  & 25                       & 0.3                     & 0.15                  & 0.64                  \\ \hline
\multirow{4}{*}{Qwen2}  & 0.5b                 & 25                       & 0.17                    & 0                     & 0.4                   \\
                        & 1.5b                 & 25                       & 0.158                   & 0                     & 0.35                  \\
                        & 7b                   & 25                       & 0.211                   & 0                     & 0.4                   \\
                        & 72b                  & 25                       & 0.219                   & 0.03                  & 1                     \\ \hline
\multirow{2}{*}{Yi}     & 6b                   & 25                       & 0.229                   & 0.033                 & 0.4                   \\
                        & 34b                  & 25                       & 0.252                   & 0.09                  & 0.4                  
\end{tabular}
\end{table}


\subsubsection{Lyrics}
As shown in Table 4, the maximum Rouge-L score for each model is 1, indicating that all these LLMs can reproduce the original lyrics. Additionally, GPT-4-turbo performed exceptionally well, significantly outperforming the other LLMs.

\begin{table}[h!]
\centering
\caption{Rouge-L results based on samples from lyrics}
\label{table4}
\begin{tabular}{c|ccccc}
\textit{\textbf{LLMs}}  & \textit{\textbf{Parameter}} & \textit{\textbf{Counts}} & \textit{\textbf{Means}} & \textit{\textbf{Min}} & \textit{\textbf{Max}} \\ \hline
\multirow{3}{*}{GPT}    & 3.5-turbo            & 371                      & 0.166                   & 0                     & 1                     \\
                        & 4-turbo              & 371                      & 0.34                    & 0                     & 1                     \\
                        & 4o                   & 371                      & 0.178                   & 0                     & 1                     \\ \hline
\multirow{3}{*}{Llama2} & 7b                   & 371                      & 0.13                    & 0                     & 1                     \\
                        & 13b                  & 371                      & 0.135                   & 0                     & 1                     \\
                        & 70b                  & 371                      & 0.194                   & 0                     & 1                     \\ \hline
\multirow{2}{*}{Llama3} & 8b                   & 371                      & 0.194                   & 0                     & 1                     \\
                        & 70b                  & 371                      & 0.276                   & 0                     & 1                     \\ \hline
\multirow{4}{*}{Qwen2}  & 0.5b                 & 371                      & 0.097                   & 0                     & 1                     \\
                        & 1.5b                 & 371                      & 0.021                   & 0                     & 0.3                   \\
                        & 7b                   & 371                      & 0.109                   & 0                     & 1                     \\
                        & 72b                  & 371                      & 0.187                   & 0                     & 1                     \\ \hline
\multirow{2}{*}{Yi}     & 6b                   & 371                      & 0.166                   & 0                     & 1                     \\
                        & 34b                  & 371                      & 0.229                   & 0                     & 1                    
\end{tabular}
\end{table}


\subsection{Analysis of Influence Factors}
Based on the above results, we observe that LLMs can output copyright protected material. Next, we will analyze the factors that influence LLMs in generating such content from the following four aspects.

\subsubsection{Parameter Scales}
Among the LLMs selected for this study, all except GPT have disclosed their parameter scales. For example, Llama2 is available in three parameter scales: 7b, 13b, and 70b, while Qwen2 offers four parameter scales: 0.5b, 1.5b, 7b, and 72b. Below, we analyze the results across different parameter scales.

We consider outputs with a Rouge-L score exceeding 0.85 to be highly similar to the original copyrighted content. Our analysis will proceed as follows: first, we will compare the mean Rouge-L scores across each parameter scale. If the means are comparable, we will then examine the frequency of results with Rouge-L scores surpassing 0.85.

\begin{table}[h!]
\begin{center}
\caption{Average Rouge-L (R-L) and Rouge-L $\geq 0.85$ results of different parameter sizes for each LLMs.}
\label{table5}
\begin{tabular}{c|c c c}
\textit{\textbf{LLMs}}  & \textit{\textbf{Parameter}} & \textit{\textbf{Mean R-L}} & \textit{\textbf{R-L $\geq \textit{\textbf{0.85}}$}} \\ \hline
\multirow{3}{*}{Llama2} & 7b                                & 0.175        & 10             \\
                        & 13b                               & 0.210        & 20             \\
                        & 70b                               & 0.244        & 43             \\ \hline
\multirow{2}{*}{Llama3} & 8b                                & 0.176        & 28             \\
                        & 70b                               & 0.360        & 101            \\ \hline
\multirow{4}{*}{Qwen2}  & 0.5b                              & 0.199        & 10             \\
                        & 1.5b                              & 0.182        & 4              \\
                        & 7b                                & 0.160        & 4              \\
                        & 72b                               & 0.194        & 32             \\ \hline
\multirow{2}{*}{Yi}     & 6b                                & 0.167        & 17             \\
                        & 34b                               & 0.228        & 46             
\end{tabular}
\end{center}
\end{table}

\begin{figure}[h!]
    \centering
    \includegraphics[width=1.0\linewidth]{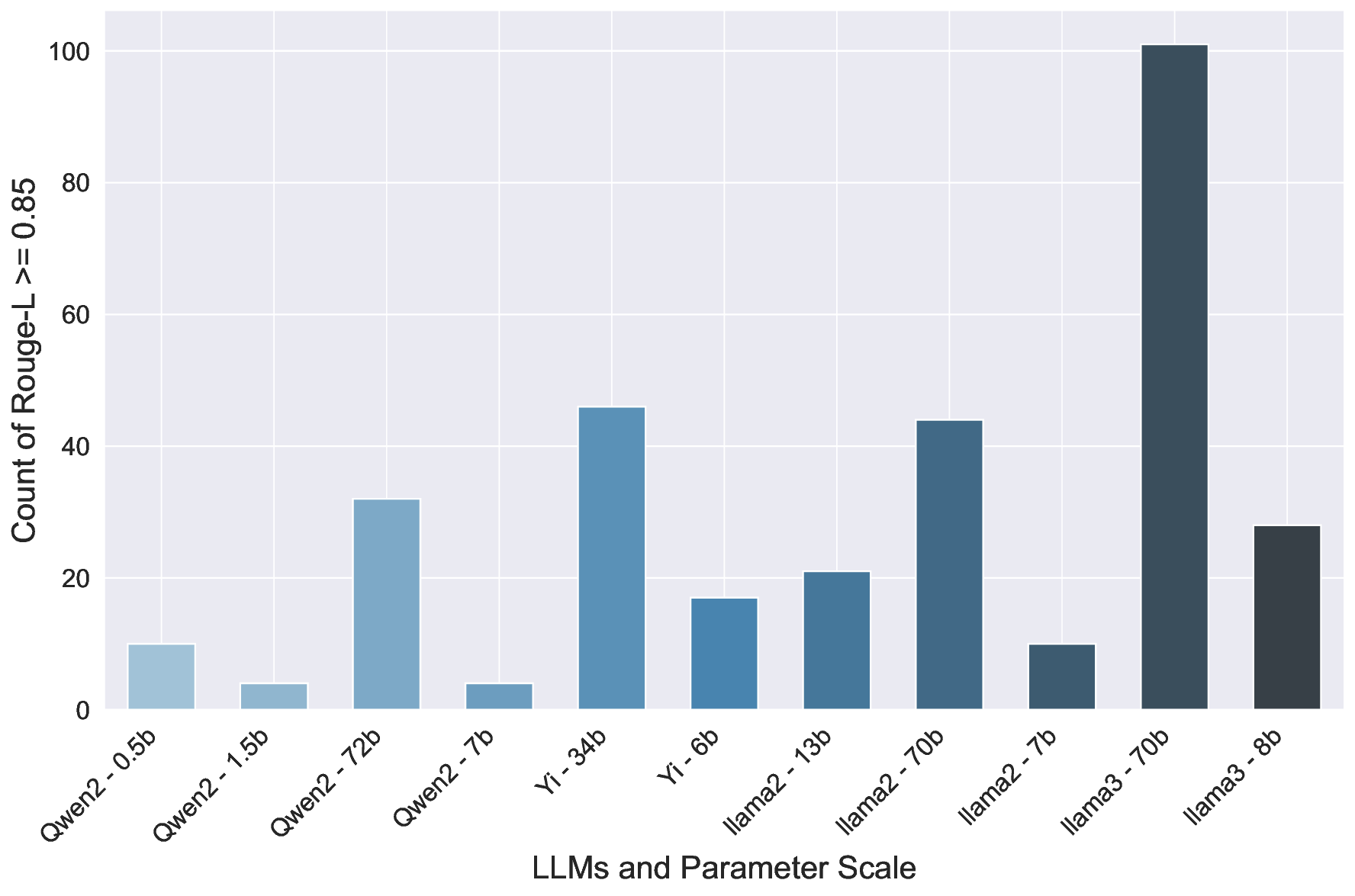}
    \caption{Count of Rouge-L Scores $\geq 0.85$ by LLMs and Parameter Scale. The x-axis represents different LLMs, while the y-axis represents the number of instances where Rouge-L is greater than or equal to 0.85. Each type of LLM is represented by a different color. }
    \Description{This figure illustrates the number of instances where the ROUGE-L score is greater than or equal to 0.85 in the test results of open-source Large Language Models (LLMs). Among these, Llama3-70b demonstrates the best performance, followed by Qwen2-72b and Yi-34b, with Llama2-70b also showing notable results. The figure provides evidence that LLMs with larger parameter scales exhibit stronger capabilities in memorizing text.}
    \label{fig:enter-label}
\end{figure}

According to Table 5 and Figure 2, we can visually observe that for Llama2, Llama3, and Yi, a larger parameter scale results in a higher similarity between the generated content and the original text. Although the average Rouge-L score for Qwen2 does not show a significant difference across scales, the number of completions with a Rouge-L score greater than 0.85 is much higher for Qwen2-72b compared to the other parameter scales. From this, we can conclude that models with larger parameter scales have better memory and are more likely to output content that closely matches the original text.

\subsubsection{Open-Source vs. Closed-Source}
Closed source LLMs typically exhibit superior performance compared to their open-source counterparts. Our objective is to analyze whether there are discernible differences between open-source LLMs (such as Llama, Qwen, and Yi) and closed-source LLMs (such as GPT).

\begin{table}[h!]
\begin{center}
\caption{Due to the varying parameter scales of different LLMs, the number of tests conducted is also different. Here, we calculate the ratio of instances where the Rouge-L (R-L) Score $ \geq 0.85$ to the total number of tests.}
\label{table6}
\begin{tabular}{c|cccc}
\textit{\textbf{LLMs}} & \textit{\textbf{Count}} & \textit{\textbf{Mean}}     & \textit{\textbf{R-L $\geq \textit{\textbf{0.85}}$}} & \textit{\textbf{Rate}}  \\ \hline
Llama2                 & 1908  & 0.184    & 73       & 0.038 \\
Llama3                 & 1272  & 0.250    & 129      & 0.101 \\
Qwen2                  & 2544  & 0.138    & 50       & 0.02  \\
Yi                     & 1272  & 0.2      & 63       & 0.049 \\
GPT                    & 1908  & 0.253778 & 192      & 0.101
\end{tabular}
\end{center}
\end{table}

As shown in Table 6, GPT models outperform most open-source models, with the exception of Llama3. Llama3, currently the most powerful open-source model, significantly outperforms its predecessor, Llama2.

Given that OpenAI has not disclosed any technical details about GPT-4, we cannot definitively determine the factors contributing to its superior ability to generate copyrighted text compared to other models. However, unconfirmed reports suggest that GPT-4 may have 1.76 trillion parameters \citep{GPT-4Architecture}, which we hypothesize could be a contributing factor.

\subsubsection{Text Type}
For this experiment, we selected three different types of texts—novels, news articles, and song lyrics—to ensure the generality of the results. All these texts are copyright protected. Based on the previous experiments, we can also analyze whether LLMs are more likely to output copyright protected material from certain types of texts.

\begin{table}[h!]
\begin{center}
\caption{Rouge-L Results of Different Text Type}
\label{table7}
\begin{tabular}{c|ccc}
\textit{\textbf{Text}} & \textit{\textbf{Count}} & \textit{\textbf{Mean}}  & \textit{\textbf{Max}}  \\ \hline
Novel                  & 3360  & 0.232 & 1    \\
News                   & 350   & 0.22  & 0.65 \\
Lyrics                 & 5194  & 0.173 & 1   
\end{tabular}
\end{center}
\end{table}

\begin{figure}[h]
  \centering
  \includegraphics[width=1\linewidth]{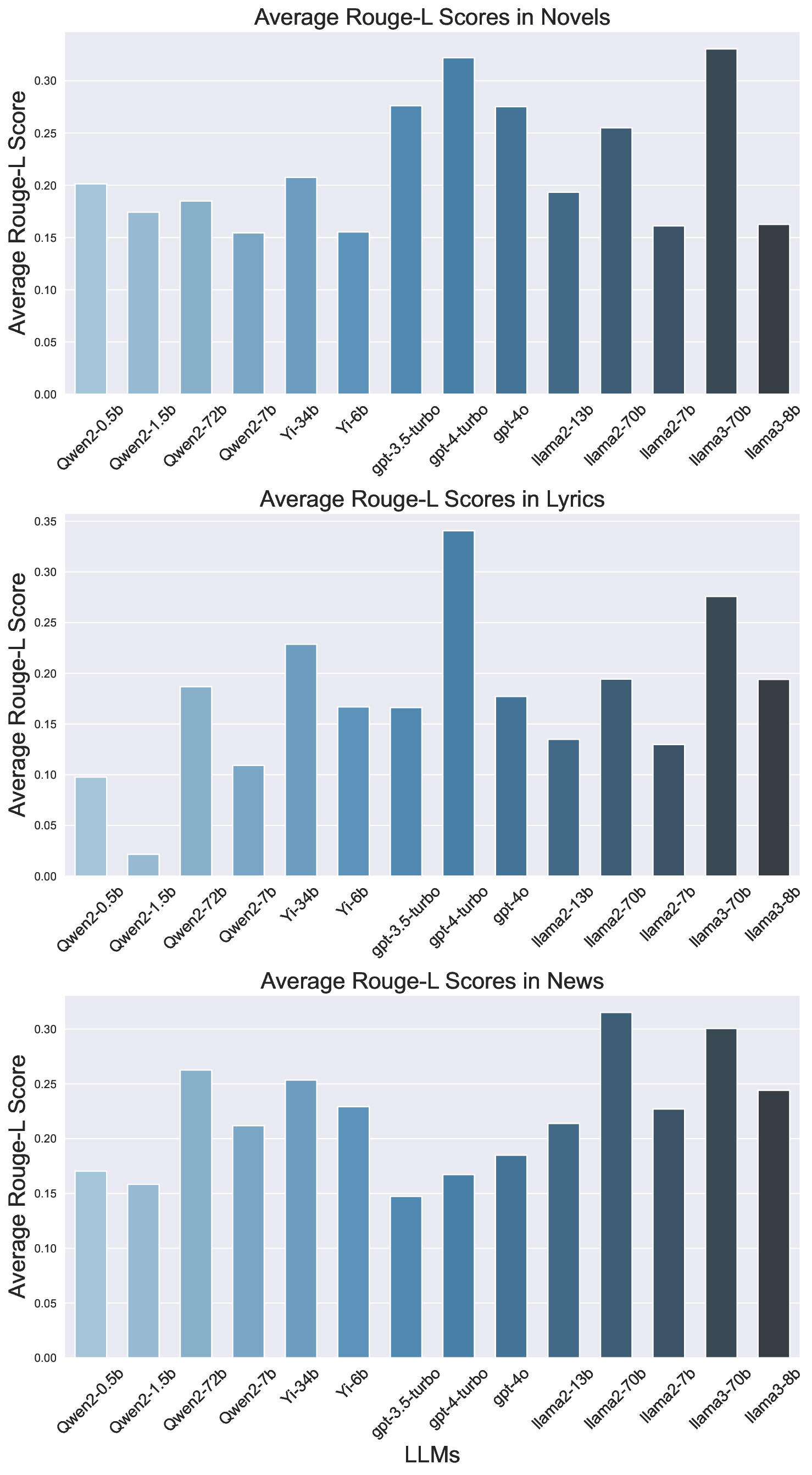}
  \caption{The average Rouge-L Scores for the three Content Types 'Novel', 'News', and 'Lyrics' can help us compare the performance of different test text types across various models. The x-axis represents different LLMs, while the y-axis represents the average Rouge-L Score for that text type.}
  \Description{This set comprises three figures, each representing the average Rouge-L scores of LLMs' completion results for different types of test texts: novels, news, and lyrics. The x-axis denotes the various LLMs, while the y-axis represents the average Rouge-L score. The first figure illustrates results for Novel texts. In this graph, Llama3-70b exhibits the highest average Rouge-L score. The second figure presents results for Lyric texts. Here, GPT-3.5-Turbo demonstrates the highest average Rouge-L score. The third figure shows results for News texts. In this case, Llama2-70b and Llama3-70b jointly display the highest average Rouge-L scores.}
  \label{fig:enter-label}
\end{figure}

According to the data in Table 7, the degree of reproduction among these three text types does not vary significantly. Next, we will examine the performance of different models across these various text types.

As shown in Figure 3, different LLMs exhibit distinct memorization capabilities for various text types. Llama3 performs well in memorizing novel content, Llama2 shows good performance with news articles, and GPT appears to favor memorizing song lyrics.

\begin{figure}[h!]
    \centering
    \includegraphics[width=1\linewidth]{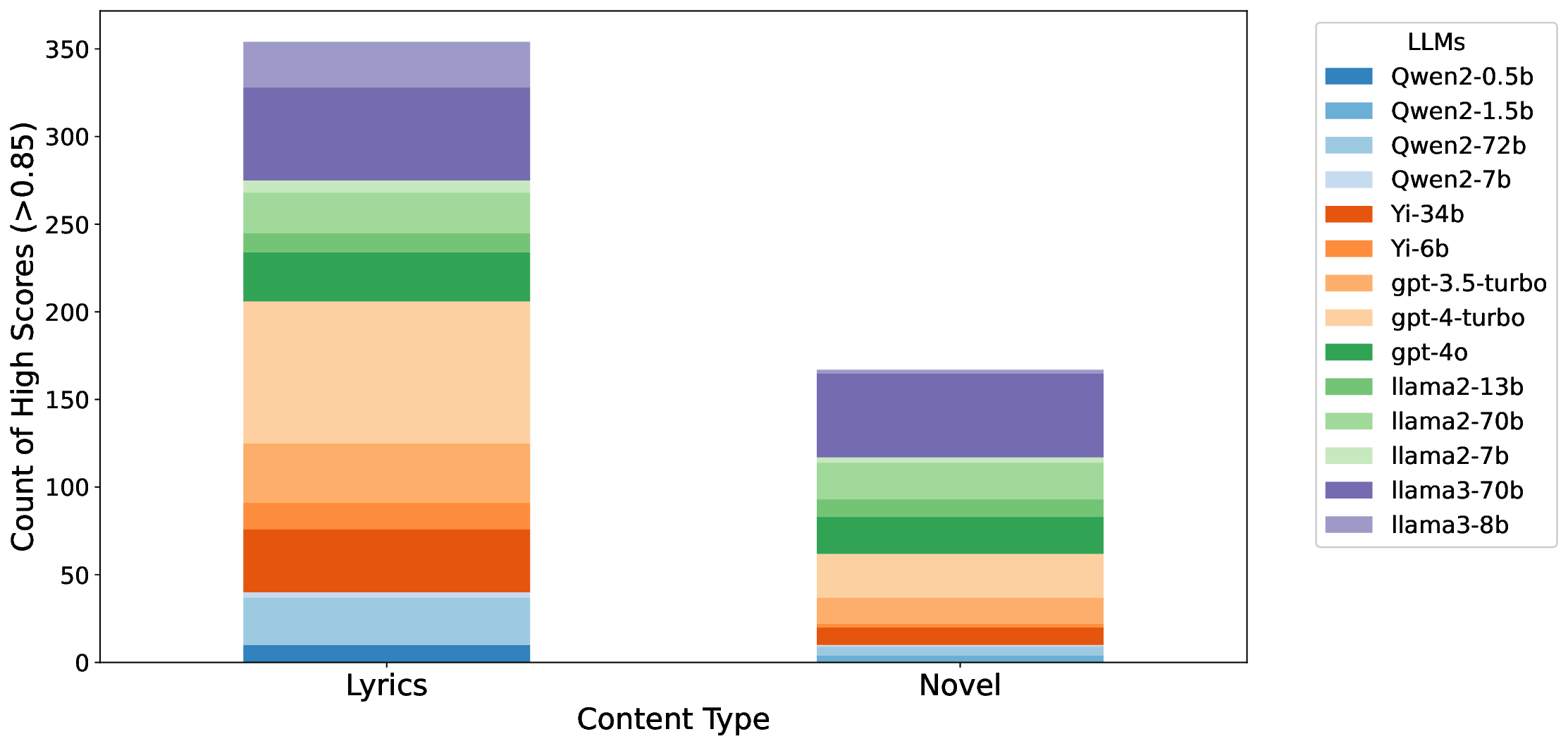}
    \caption{Rouge-L $ \geq 0.85$ for Novels and Lyrics. The x-axis represents different LLMs, while the y-axis represents the number of instances where Rouge-L is greater than or equal to 0.85. The size of different colors in the stacked chart represents the number of instances where Rouge-L $ \geq 0.85$.}
    \Description{This figure presents a stacked bar chart. The x-axis represents the types of test texts, categorized into two groups: novels and lyrics. The y-axis indicates the number of instances where the Rouge-L score is greater than or equal to 0.85 in the test results. For the Novels category, Llama3-70b demonstrates the best performance. In the Lyrics category, GPT-4-Turbo exhibits the highest performance.}
    \label{fig:enter-label}
\end{figure}

Since there were no instances of completions with a Rouge-L score greater than 0.85 for the news articles, we will focus our analysis on novels and song lyrics. As shown in Figure 3, Llama3 excels in memorizing novel content, while GPT performs well in memorizing song lyrics, which is consistent with the results in Figure 4.

\subsubsection{Different Output Lengths}
We also tested whether the length of the output affects the content generated by LLMs. In the LLM settings, max\_tokens specifies the maximum length of the output content. Therefore, we tested the Rouge-L Score for the LLMs with max\_tokens set to 50, 100, 200 and 300. Due to the length limitations of news articles and song lyrics, we only tested the LLMs' output for novel content. To control variables, the length of the input prompt in all tests in this section was consistently 50 tokens.

\begin{table}[h!]
\begin{center}
\caption{Average Rouge-L (R-L) and Counts of Rouge-L $\geq 0.85$ for Different Max\_tokens}
\label{table7}
\begin{tabular}{c|cccc}
\textit{\textbf{max\_tokens}} & \textit{\textbf{Count}} & \textit{\textbf{Mean}}  & \textit{\textbf{Max}} & \textit{\textbf{R-L $\geq \textit{\textbf{0.85}}$}} \\ \hline
50                            & 1120  & 0.227 & 1 & 67   \\
100                           & 1120  & 0.226 & 1 & 48   \\
200                           & 1120  & 0.242 & 1 & 38   \\
300                           & 1120  & 0.175 & 1 & 14
\end{tabular}
\end{center}
\end{table}

\begin{figure}[h!]
    \centering
    \includegraphics[width=1\linewidth]{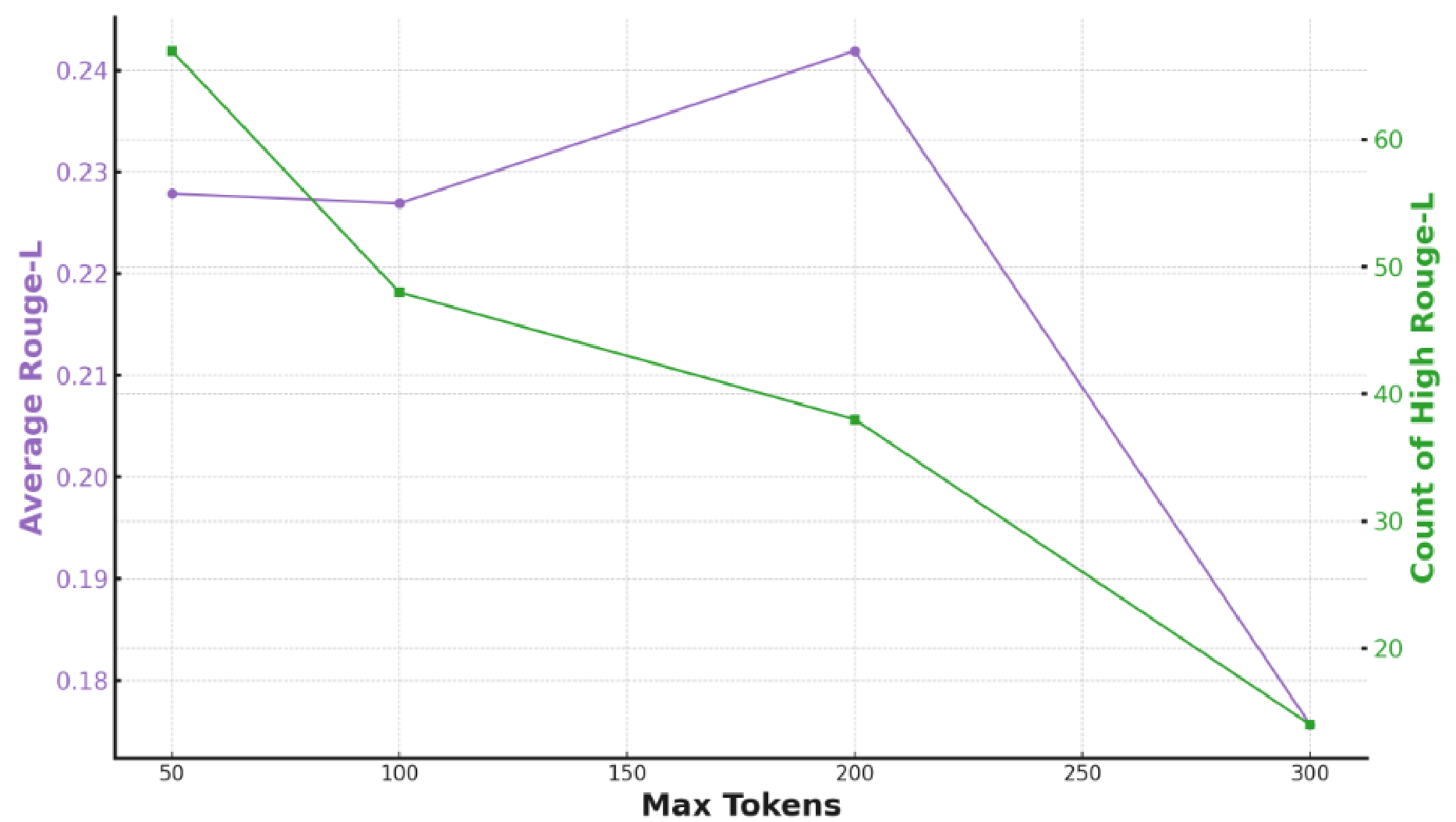}
    \caption{Comparison of Average Rouge-L and Count of High Scores by Max Tokens. The x-axis represents the size of max\_tokens. The left y-axis represents the average Rouge-L score, while the right y-axis represents the number of instances where Rouge-L $ \geq 0.85$. The average Rouge-L is represented in red, and the number of instances where Rouge-L $ \geq 0.85$ is represented in blue.}
    \Description{Comparison of Average Rouge-L and Count of High Scores by Max Tokens. The x-axis represents the size of max\_tokens. The left y-axis represents the average Rouge-L score, while the right y-axis represents the number of instances where Rouge-L $ \geq 0.85$. The average Rouge-L is represented in red, and the number of instances where Rouge-L $ \geq 0.85$ is represented in blue. The figure demonstrates that the average Rouge-L score initially increases as max_tokens reaches 200, at which point it attains its maximum value. Subsequently, the average Rouge-L score begins to decline. Conversely, the Count of High Rouge-L scores exhibits a decreasing trend as max_tokens increases.}
    \label{fig:enter-label}
\end{figure}

Table 8 shows the average Rouge-L Scores and the number of scores above 0.85 for different max\_tokens. To facilitate observation, we also created Figure 5. As shown in Figure 5, the average Rouge-L Score unexpectedly showed a brief upward trend as max\_tokens increased, which contradicts our inference. However, when we observe the high-scoring Rouge-L, we find that the number of high scores indeed decreases as max\_tokens increases. Therefore, we will analyze the performance of each LLM under different max\_tokens to seek the reason for this phenomenon.

\begin{figure}[t]
    \centering
    \includegraphics[width=1\linewidth]{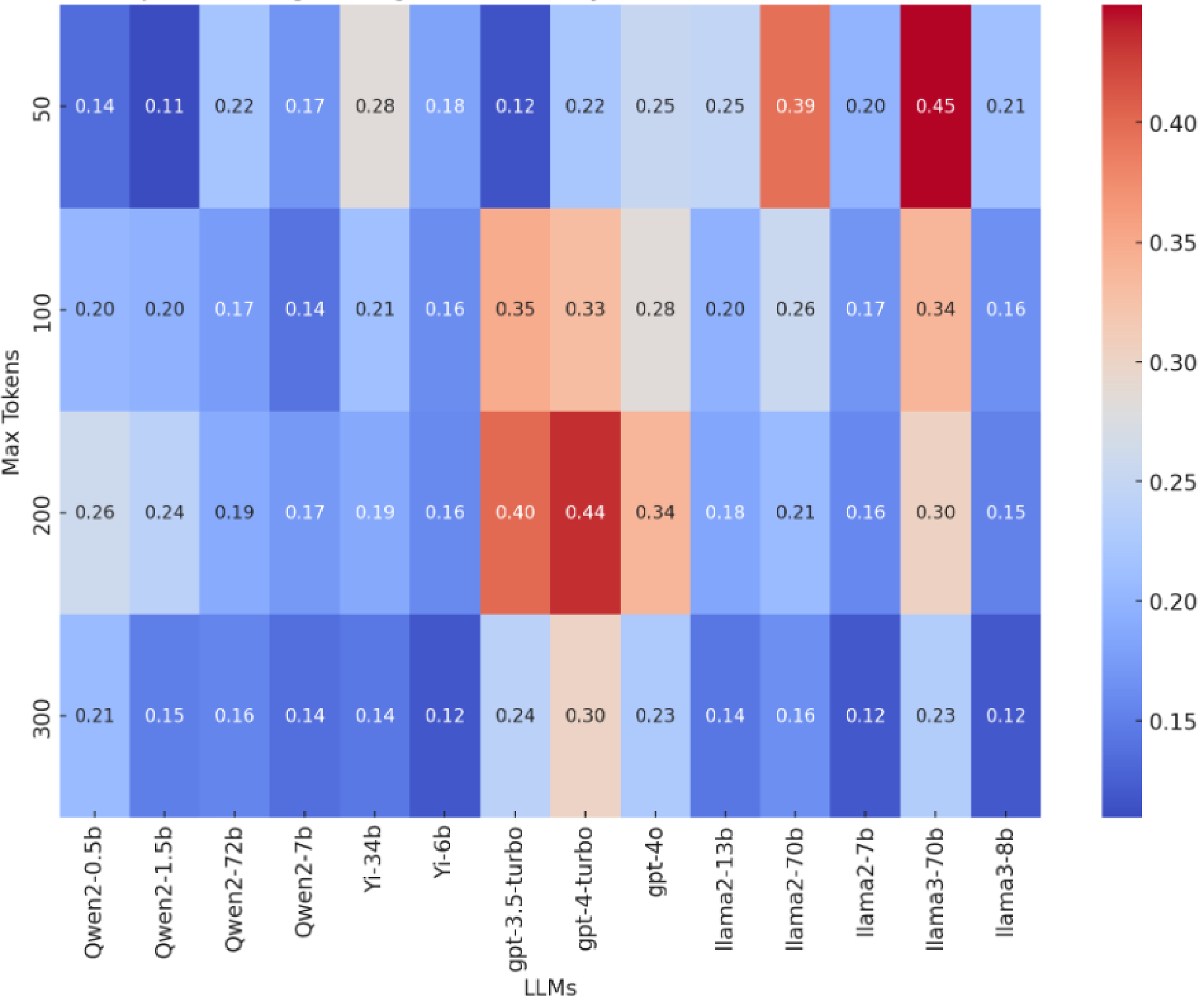}
    \caption{Heatmap for Average Rouge-L by Max\_tokens and LLMs. The x-axis represents different LLMs, while the y-axis represents different max\_tokens values. The values in the heatmap indicate the average Rouge-L score. The redder the color, the higher the average Rouge-L score; the bluer the color, the lower the Rouge-L score.}
    \Description{Heatmap for Average R-L by Max\_tokens and LLMs. The x-axis represents different LLMs, while the y-axis represents different max\_tokens values. The values in the heatmap indicate the average Rouge-L score. The redder the color, the higher the average Rouge-L score; the bluer the color, the lower the Rouge-L score. The figure illustrates that GPT exhibits its optimal performance when max_tokens is set to 200. In contrast, for all other Large Language Models (LLMs) represented, the average Rouge-L score demonstrates a decreasing trend as max_tokens increases.}
    \label{fig:enter-label}
\end{figure}

As shown in Figure 6, we can observe that the average Rouge-L scores of all open-source LLMs decrease as max\_tokens increases. However, for the closed-source LLM GPT, the average Rouge-L score reaches its peak when max\_tokens is 200, and then begins to decrease. For now, let's set aside the discussion of GPT and examine the Rouge-L scores of all open-source LLMs, before finally analyzing why GPT differs from other LLMs.

\begin{figure}
    \centering
    \includegraphics[width=1\linewidth]{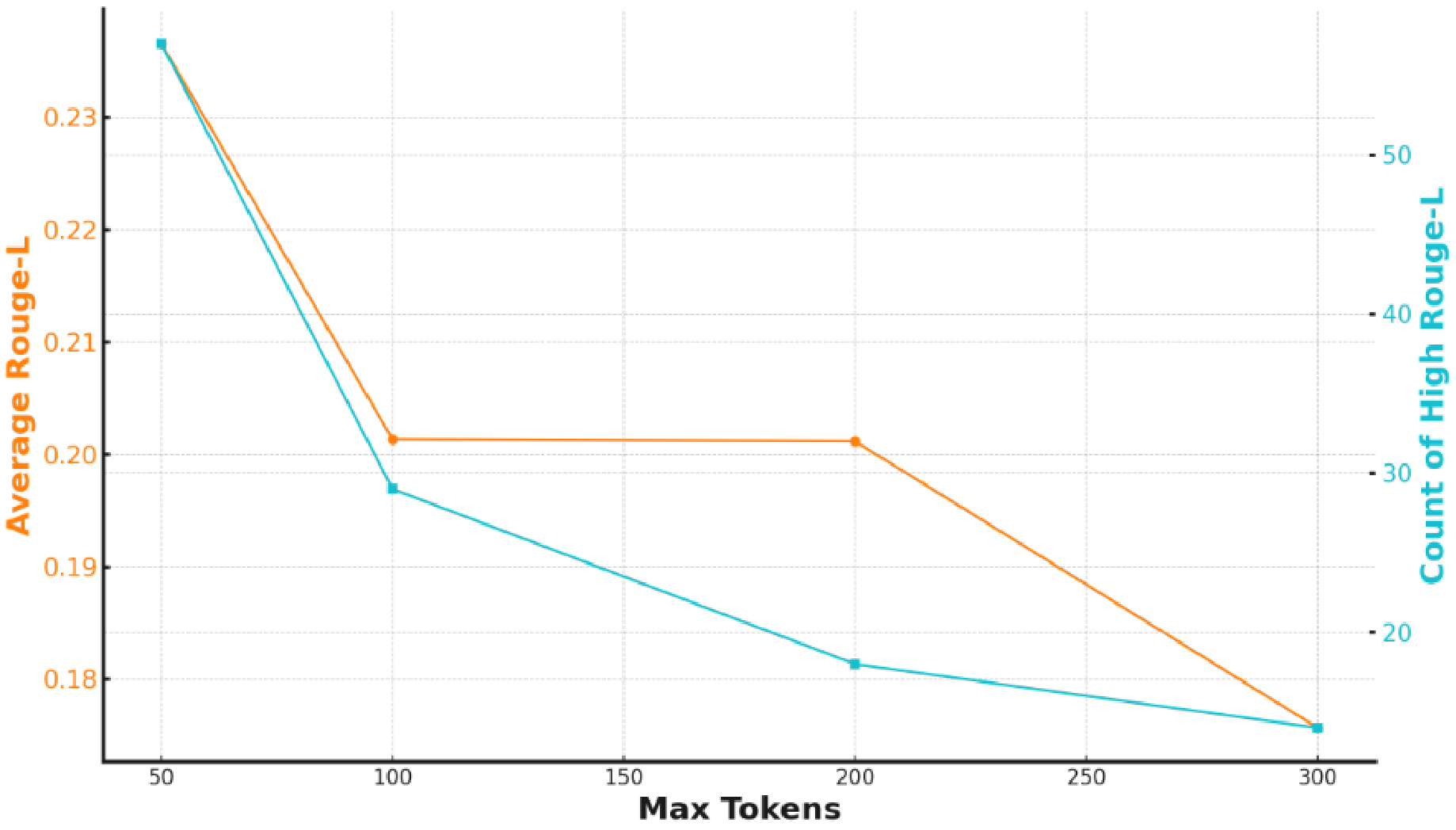}
    \caption{Comparison of Average Rouge-L and Count of High Scores by Max Tokens (without GPT). The x-axis represents the size of max\_tokens. The left y-axis represents the average Rouge-L score, while the right y-axis represents the number of instances where Rouge-L $ \geq 0.85$. The average Rouge-L is represented in red, and the number of instances where Rouge-L $ \geq 0.85$ is represented in blue.}
    \Description{The x-axis represents the size of max\_tokens. The left y-axis represents the average Rouge-L score, while the right y-axis represents the number of instances where Rouge-L $ \geq 0.85$. The average Rouge-L is represented in red, and the number of instances where Rouge-L $ \geq 0.85$ is represented in blue. The figure demonstrates that both the average Rouge-L score and the Count of High Rouge-L decrease as max_tokens increases.}
    \label{fig:enter-label}
\end{figure}

According to Figure 7, we can observe that as max\_tokens increases, the text reproduction ability of all open-source models decreases with the increase of max\_tokens.

We believe there are two possible reasons for this phenomenon. The first reason is the randomness of LLMs' output. Even when we set the temperature to 0 or 0.01, it does not guarantee that the LLMs' output will be deterministic. According to the output logs, LLMs generate their output in segments, with each segment matching the most likely result from their database. This means that as the max\_tokens value increases, even if the initial part of the output is highly similar to the original text, the later segments may diverge significantly. So in partial results, the first half of the LLMs completed content overlapped very high, but when the output reached a certain length, it suddenly began to output completely unrelated content.

The second possible reason is that different LLMs have different safety measures. It is well-known that LLMs should not output copyright protected content. To prevent this, developers implement various safety measures for their LLMs. For example, LLMs like ChatGPT or Claude, which are operated online, have real-time content detection safety measures. These LLMs continuously monitor their output to detect and prevent potentially illegal or copyright infringing content.

\subsection{Iterative Prompting Result}
We utilized the first, second, and third books of the "\emph{Harry Potter}" series as test materials. The initial sentence of each book was provided to the LLMs as a prompt, followed by iterative prompting, wherein each completion generated by the LLMs was used as a new prompt. The results are illustrated in Figure 8. Our findings indicate that this method indeed enables LLMs to generate more copyrighted content. However, after a certain number of iterations, the LLMs began to produce content entirely unrelated to the copyrighted material. We hypothesize that this deviation may occur due to the LLMs generating content divergent from the original material in one iteration, consequently leading to low overlap in subsequent completions.
\begin{figure}[h]
    \centering
    \includegraphics[width=1\linewidth]{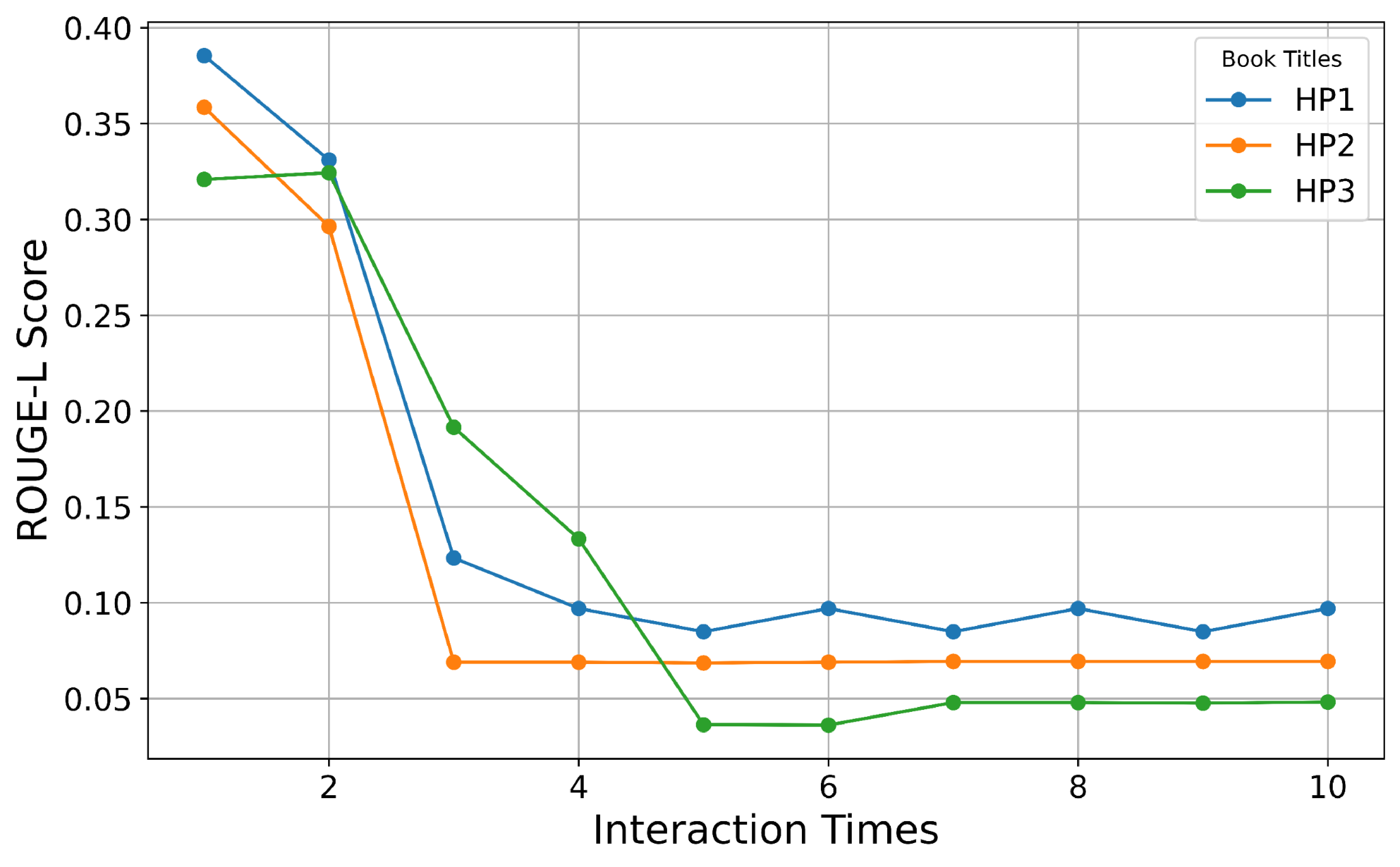}
    \caption{Rouge-L Scores Across Iterations for Different Books.The X-axis represents the number of iterations, and the Y-axis represents the Rouge-L Score of each LLMs output. The three curves represent "\emph{Harry Potter}" 1, 2, and 3.}
    \label{fig:enter-label}
\end{figure}

\subsection{Summary}
In our experiment, the LLMs successfully reproduced text highly similar to the copyrighted material based on the partial content we provided. Models with larger parameter sizes demonstrated a stronger ability to reproduce the copyrighted material. Different text types and variations among LLMs also influenced the final results. Finally, we discovered that each LLM has a certain threshold: when the maximum output text length exceeds this threshold, its reproduction ability significantly weakens.

\section{Conclusion}
Our experiment simulated a scenario where individuals obtain a small segment of a copyright protected work and then use current LLMs to generate the subsequent content based on this segment. The results demonstrate that LLMs are capable of generating copyright infringing materials. We identified four factors that influence LLMs' propensity to output infringing content, with the scale of parameters and maximum output length having a significant impact. Additionally, we found that iterative prompting enables LLMs to generate more content that could potentially constitute copyright infringement. However, after a certain number of iterations (for instance, the third output in our experiment), the LLMs began to produce content entirely dissimilar to the original material.



\section{Limitations and Future Research Directions}
The datasets employed in this study primarily focus on English-language texts, which may not accurately reflect the behavior of LLMs with copyrighted texts in other languages or other types of materials (such as code). Furthermore, our exclusive use of Rouge-L as the evaluation metric may lead to misclassification for certain samples. Although this study utilized multiple LLMs, numerous recently released models remain untested. These newer models may possess more robust defense mechanisms, potentially reducing their likelihood of generating infringing content. We also experimented with a method that allows LLMs to self-iterate in generating prompts. While this approach led to the production of more infringing content, there remains substantial room for improvement. Potential refinements could include adjusting the maximum output length for iterations to avoid triggering LLMs' safety measures, as well as exploring the use of more powerful models.

We propose three directions for future research. First, developing suitable algorithms using LLMs with publicly available training data and details would allow researchers to input a small segment of content and generate extensive subsequent content. Second, examining potential differences between texts in various languages is crucial. For example, comparing English and Chinese texts is important since Chinese uses a completely different tokenizer from English, implying significant differences in training details. To ensure the legality of LLM outputs, special attention must be paid to textual content across different languages. Third, in relation to the Iterative Prompting approach, one could explore the application of jailbreak attacks on LLMs to circumvent their safety protocols, potentially inducing the generation of more infringing content.



\bibliographystyle{ACM-Reference-Format}
\bibliography{References}

\appendix

\end{document}